\documentclass[10pt,twocolumn,letterpaper]{article}

\usepackage{iccv}
\usepackage{times}
\usepackage{epsfig}
\usepackage{graphicx}
\usepackage{amsmath}
\usepackage{amssymb}
\usepackage{multirow}
\usepackage{tabularx}
\usepackage{booktabs}
\usepackage{ulem}
\usepackage{threeparttable}
\usepackage[breaklinks=true,bookmarks=false]{hyperref}

\iccvfinalcopy % *** Uncomment this line for the final submission

\def\iccvPaperID{****} % *** Enter the ICCV Paper ID here

\ificcvfinal\fi

\begin{document}

%%%%%%%%% TITLE
\title{Part-Aware Transformer for Generalizable Person Re-identification}

\author{Hao Ni\textsuperscript{1}\hspace{-0.2cm}\and
Yuke Li\textsuperscript{1}\hspace{-0.2cm}\and
Lianli Gao\textsuperscript{2}\hspace{-0.2cm}\and
Heng Tao Shen\textsuperscript{1}\hspace{-0.2cm}\and
Jingkuan Song\textsuperscript{2}\thanks{Jingkuan Song is the corresponding author.} \and 
\textsuperscript{1} University of Electronic Science and Technology of China (UESTC)
\\
\textsuperscript{2}Shenzhen Institute for Advanced Study, UESTC \\
{\tt\small {\{haoni0812, liyuke65535, jingkuan.song\}@gmail.com}, 
}
%\href{mailto:liyuke65535@gmail.com}{liyuke65535@gmail.com}, \href{mailto:jingkuan.song@gmail.com}{jingkuan.song@gmail.com}
}

\maketitle
\ificcvfinal\thispagestyle{empty}\fi

%%%%%%%%% ABSTRACT
\begin{abstract}
   Domain generalization person re-identification (DG-ReID) aims to train a model on source domains and generalize well on unseen domains.
   Vision Transformer usually yields better generalization ability than common CNN networks under distribution shifts. 
   However, Transformer-based ReID models inevitably over-fit to domain-specific biases due to the supervised learning strategy on the source domain.
   We observe that while the global images of different IDs should have different features, their similar local parts (e.g., black backpack) are not bounded by this constraint. 
   Motivated by this, we propose a pure Transformer model (termed Part-aware Transformer) for DG-ReID by designing a proxy task, named Cross-ID Similarity Learning (CSL), to mine local visual information shared by different IDs. This proxy task allows the model to learn generic features because it only cares about the visual similarity of the parts regardless of the ID labels, thus alleviating the side effect of domain-specific biases. 
   Based on the local similarity obtained in CSL, a Part-guided Self-Distillation (PSD) is proposed to further improve the generalization of global features. 
   Our method achieves state-of-the-art performance under most DG ReID settings. Under the Market$\to$Duke setting, our method exceeds state-of-the-art by 10.9\% and 12.8\% in Rank1 and mAP, respectively.
   % Anonymous code is available at \url{https://anonymous.4open.science/r/Part-Aware-Transformer}.
   The code is available at \url{https://github.com/liyuke65535/Part-Aware-Transformer}.
\end{abstract}

%%%%%%%%% BODY TEXT
\section{Introduction}
\label{sec:intro}

Person Re-Identification (ReID)~\cite{reid0,reid1,reid2,reid3} aims to find persons with the same identity from multiple disjoint cameras. Thanks to the great success of Convolutional Neural Network~(CNN)~ in the field of computer vision \cite{CNN1, CNN2}, supervised, unsupervised person ReID has made significant progress. However, a more challenging task, domain generalization (DG) ReID~\cite{general1} which trains a model on source domains yet generalizes well on unseen target domains, still lags far behind the performance of supervised ReID model.

Thus, many DG methods are proposed to learn generic features. These methods explore the generalization of CNN based on disentanglement~\cite{SNR} or meta-learning~\cite{metabin, MDA}. Recently, Transformer has gained increasing attention in computer vision. It is a neural network based on attention mechanisms~\cite{transformer}. Vision Transformer usually yields better generalization ability than common CNN networks under distribution shift\cite{DG_Transformer}. However, existing pure transformer-based ReID models are only used in supervised and pre-trained ReID~\cite{TransReID_pretrain,transreid}. The generalization of Transformer is still unknown in DG ReID. 

%Designed for natural language processing (NLP), the Transformer is notable for long-range sequence data. Researchers investigated its adaptation to computer vision. It has also shown promising performance for computer vision tasks, including image classification object detection\cite{trans_obj1, trans_obj2}, and image segmentation\cite{trans_seg1, trans_seg2}.

\begin{figure}[t]
    \begin{center}
        \includegraphics[width=1\columnwidth]{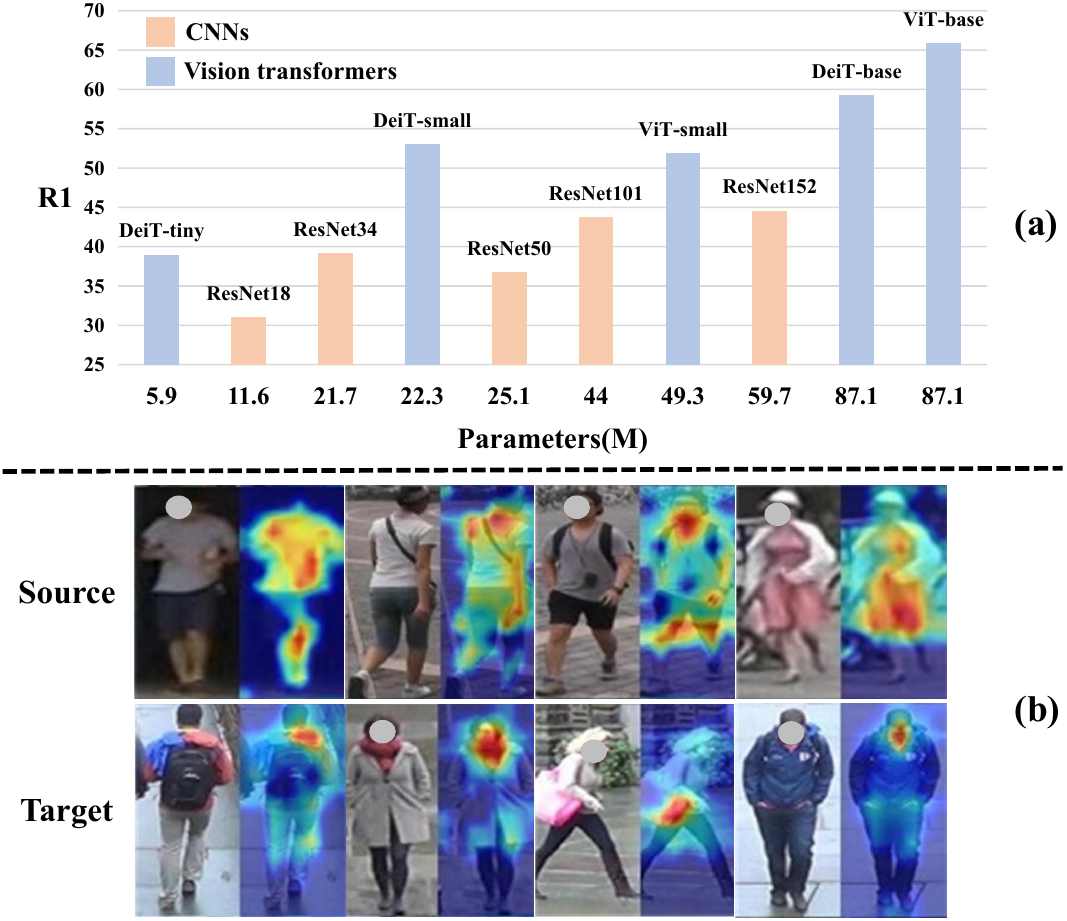}
    \end{center}
    \setlength{\abovecaptionskip}{0.cm}
\caption{(a) We applied different Transformers to DG ReID. Models are trained on Market and tested on Duke. Results show Vision transformers (blue bars) are better than CNNs (orange bars) even with fewer parameters. (b) Visualization of attention maps of ``class token" on source domain (Market) and target domain (Duke). We use ViT~\cite{vit} as the backbone and fuse the attention results of the shallow layers. However, the attention to discriminative information is still limited on target domain.}
    \label{fig:intro_2}
\end{figure}

To investigate the performance of Transformer in DG ReID, we use different Transformers and CNNs as backbones to test their cross-domain performance from Market to Duke. The results show that Vision Transformers are much better than common CNNs, as shown in Figure~\ref{fig:intro_2} (a). Even with fewer parameters, Transformers still outperform CNNs. For instance, DeiT-tiny \cite{DeiT} with 5.9M parameters is much better than ResNet18 with 11.6M parameters. Despite the great performance of ViT~\cite{vit}, we still experimentally find that the attention to discriminative information is limited on unseen target domain. As shown in Figure~\ref{fig:intro_2}(b), some discriminative information on the target domain is ignored, such as the black backpack and grey coat. 

\begin{figure}
    \begin{center}
        \includegraphics[width=1\columnwidth]{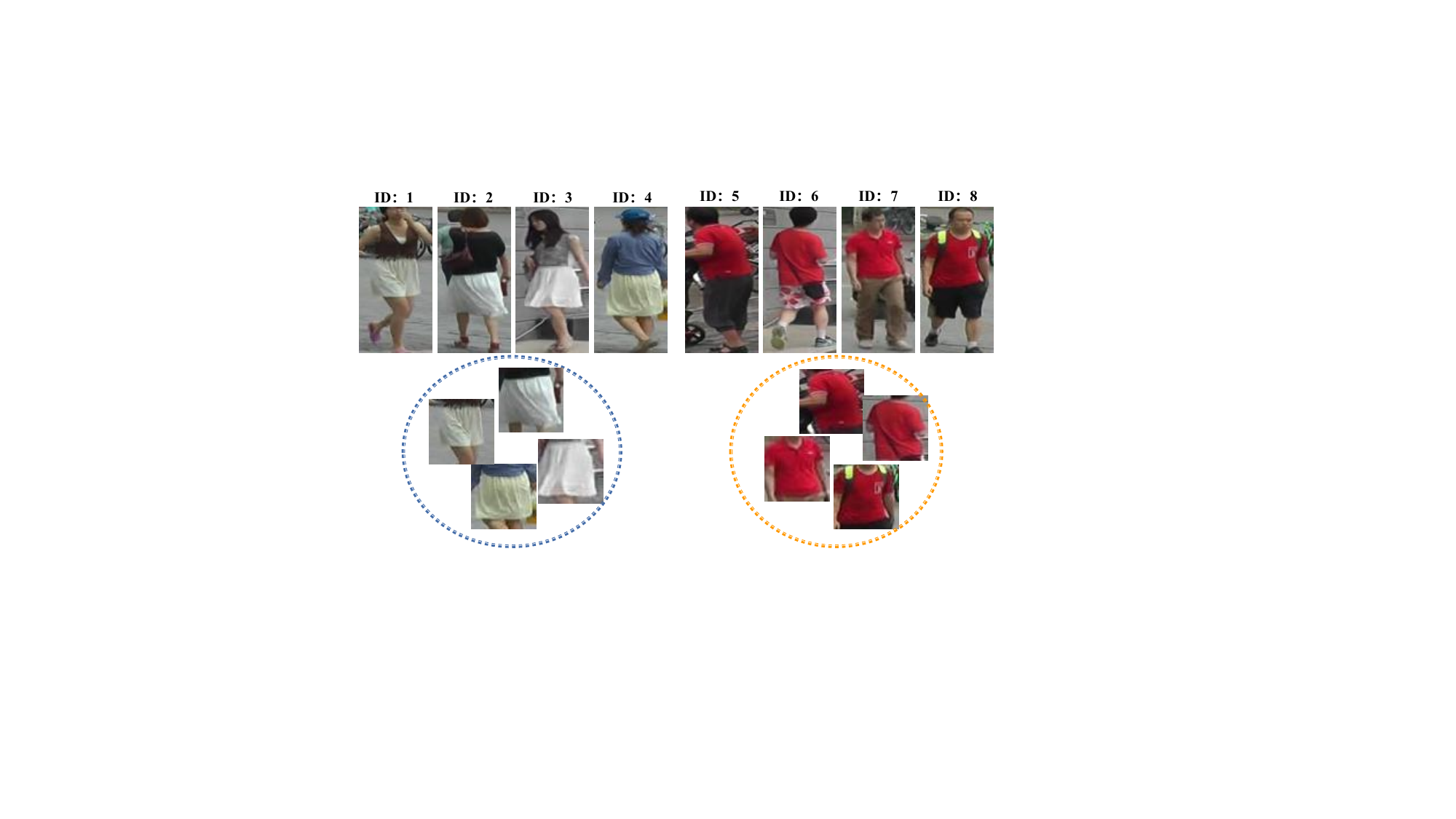}
    \end{center}
    \setlength{\abovecaptionskip}{0.cm}
    \caption{Local similarity among parts with different IDs. It comes from the visual data themselves, not from ID labels.}
    %We propose CSL and PSD to learn generic and generalized features through this apparent similarity. }
    \label{fig:intro_1}
\end{figure}

The above phenomenon shows that Transformer-
based ReID models inevitably over-fit to domain-specific biases due to the supervised learning strategy on the source domain. It is manifested in the insufficient learning of local information on unseen target domains. We observe that while the global images of different IDs should have different features, their similar local parts (e.g., White skirt, red T-shirt) are not bounded by this constraint, as shown in Figure \ref{fig:intro_1}. And these ID-independent local similarities can provide extra visual knowledge from the images themselves. Ignoring this similarity leads the Transformer to focus on the ReID task instead of learning generic features, resulting in more over-fitting to domain-specific biases. 

To this end, we design a proxy task, named Cross-ID Similarity Learning (CSL), to mine local similarities shared by different IDs and learn generic features without using ID labels. CSL is based on part-aware attention to learn discriminative information across different IDs. The part-aware attention concatenates the ``part token" and the ``image tokens" in the region of interest to learn local representations. In each mini-batch, we use a memory bank to calculate the distance between the current local features and the samples of the entire dataset to mine apparent local similarity. The apparent similarity is learned not from ID annotations, but from the visual data themselves~\cite{instance_dis}. Thus it allows the model to learn generic features because it only cares about the visual similarity of the parts regardless of the ID labels, thus alleviating the side effect of domain-specific biases.

%The motivations for designing part-aware attention and CSL are as follows: 1) The inductive bias of the Transformer is less than CNN, such as translation invariance. Therefore, it requires a larger amount of training data to obtain generalized representations. Part-aware attention aims to add such an inductive bias to Transformer: human images are structured, and We can identify a person's ID by comparing different parts of the person, such as the head, upper body, and lower body. %This inductive bias is only applicable to ReID and cannot be extended to general image classification. 
%2) Since the source dataset is usually small in DG setting, the model is likely to get the correct ID by mining limited information. For example, we can discriminate ID based on the mask if there is only one person wearing a mask. That is, the model only needs to focus on a limited area to make the loss converge. CSL forces the model to focus on more areas with discriminative information shared by different IDs. Besides, solving ID-independent tasks allows the Transformer to learn generic features, which reduces over-fitting to domain-specific biases.
%This process enriches the information of features and prevents the model from discriminating the ID through one-sided information.

Part-guided Self-Distillation (PSD) is proposed to further improve the generalization of the global representation. Self-distillation has been proven effective in DG image classification~\cite{DG_SD, DG_SD2}. It can learn visual similarities beyond hard labels and make the model converges easier to the flat minima. However, we experimentally find that the traditional self-distillation method would reduce the generalization in DG ReID. The reason is that ReID is a fine-grained retrieval task, and the difference between different categories is not significant. It is difficult to mine useful information from the classification results. 
Therefore, PSD uses the results of CSL to construct soft labels for global representation. In general, the motivation of self-distillation is similar to CSL, which is to learn generic features by data themselves regardless of the ID labels. 

Extensive experiments have proved that CSL and PSD can improve the generalization of the model. Specifically, our method achieves state-of-the-art performance under most DG ReID settings, especially when using small source datasets. Under the Market→Duke setting, our method exceeds state-of-the-art by 10.9\% and 12.8\% in Rank1 and mAP, respectively. The contributions of this work are three-fold:

a) We propose a pure Transformer-based framework for DG ReID for the first time. Specifically, we design a proxy task, named Cross-ID Similarity Learning module (CSL), to learn generic features. 

b) We design part-guided self-distillation (PSD) for DG ReID, which learns visual similarities beyond hard labels to further improve the generalization. 
%It constructs soft labels according to local similarity instead of the classification result.
%, which solves the problem that the difference between different IDs is too small.

c) Extensive experiments have proved that our Part-aware Transformer achieves state-of-the-art of DG ReID.
%-------------------------------------------------------------------------

\section{Related Work}
\subsection{Domain Generalizable Person ReID.}
Supervised and unsupervised domain adaptation person ReID have achieved great success. But DG ReID is still a challenging task. It requires the model to train a model on source domains yet generalize to unseen target domains. Due to its huge practical value, it has been widely studied in recent years. The concept of DG ReID was first proposed in~\cite{general1}. \cite{benchmark, ibn2} applied meta-learning to learn domain-invariant features. \cite{SNR} proposed to disentangle identity-irrelevant information. Last but not the least, \cite{BINnet} proposed IBN-net to explore the effect of combining instance and batch normalization, which was widely used in later DG ReID methods due to its good transferability and effectiveness. However, pure Transformer does use batch normalization, so IBN cannot bring gain to our model. But even without using IBN, our method still outperforms the existing CNN-based state-of-the-art in DG ReID.

\subsection{Transformer-related Person ReID.}
The original Transformer is proposed in ~\cite{transformer} for natural language processing (NLP) tasks. Based on ViT~\cite{vit}, ~\cite{transreid} applies pure Transformer to supervised ReID for the first time, which introduces side information to improve the robustness of features. ~\cite{TransReID_pretrain} further proposed self-supervised pre-training for Transformer-based person ReID, which mitigates the gap between the pre-training and ReID datasets from the perspective of data and model structure. 

Recently, some work investigate the generalization of vision Transformers~\cite{DG_Transformer}. In DG ReID, TransMatcher~\cite{transmatcher} employs hard attention to cross-matching similarity computing, which is more efficient for image matching. However, it still uses CNN as the main feature extractor, and the role of the Transformer is mainly reflected in image matching. Our method is the first to investigate the generalization ability of pure Transformer in DG ReID.

\subsection{Proxy Task and Self-Distillation.}
Proxy Task and Self-Distillation have been extensively studied, and we only discuss their contribution to generalization here.

\textbf{Proxy Task} is referred to as learning with free labels generated from the data itself, such as solving Jigsaw puzzles~\cite{Jigsaw_puzzles}, predicting rotations~\cite{predicting_rotations} or reconstruction~\cite{DG_reconstruction}. Since these tasks are not related to the target task (such as image classification), they can guide the model to learn generic features, which leads to less over-fitting to domain-specific biases~\cite{DG_self}. 
%Our CSL is closer to recent state-of-the-art SSL methods~\cite{moco,bootstrap}, which are mostly based on combining contrastive learning with data augmentation. 
CSL picks similar parts from the entire dataset without using ID labels, thus encouraging the model to learn the discriminative information shared by different IDs. 

\textbf{Self-Distillation} (SD) uses soft labels containing ``richer dark knowledge", which can reduce the difficulty of learning the mapping and further improve the generalization ability of the model~\cite{DG_SD}. Besides, ~\cite{DG_SD2} found that SD can help models converge to flat minima, improving the generalization of features. However, traditional SD methods are not suitable for ReID. Because it is a fine-grained retrieval task. So we propose PSD to replace traditional methods.

%-------------------------------------------------------------------------
\section{Methodology}

\begin{figure*}[t]
    \begin{center}
        \includegraphics[width=0.88\textwidth]{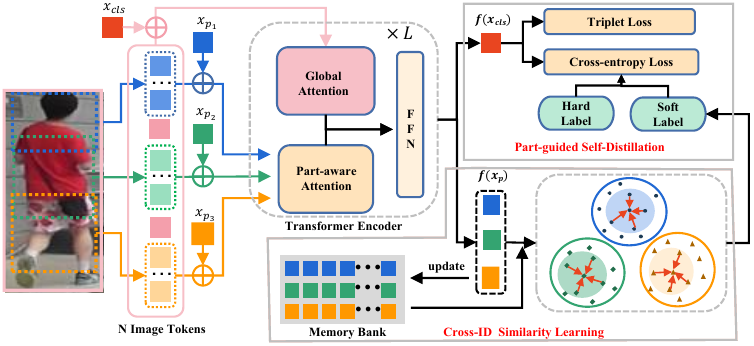}
    \end{center}
    \setlength{\abovecaptionskip}{0.cm}
    \caption{Illustration of our proposed method. FFN is a feed-forward network. We split an input image into $N$ non-overlapping patches as ``image tokens". The input of the model includes a ``class token" ($x_{cls}$), three ``part tokens" ($\{x_{p_i}| i \in {1,2,3}\}$) and $N$ ``image tokens". The ``class token" concatenates all ``image tokens" to get a global feature through global attention. For each ``part token", it concatenates ``image tokens" of the region of interest to obtain the local feature belonging to this region. The output of the $L$ Transformer block includes a global feature $f(x_{cls})$ and three local features $\{f(x_{p_i}) | i \in {1,2,3}\}$.
    We use three local representations and samples in the memory bank to solve a proxy task, named Cross-ID Similarity Learning (CSL). The results of the CSL guide the global representation to perform Part-guided Self-Distillation (PSD). }
    \label{fig:framework}
\end{figure*}
We proposed a pure Transformer-based framework, named Part-aware Transformer (PAT), to learn generalizable features, as shown in Figure~\ref{fig:framework}. In the following, we describe the main components of our method. First, we introduce our Transformer encoder composed of $L$ blocks, which simultaneously extracts global and local features (Sec. \ref{sec:Part-ware Attention}). Next, we design a proxy task, named Cross-ID Similarity Learning (CSL), to learn the generic features (\ref{sec:CSL}). It mines local similarity shared by different IDs and encourages model to learn generic features, thereby reducing over-fitting on source datasets. Finally, a Part-guided Self-Distillation module (PSD) is proposed to further improve the generalization of global features (Sec.~\ref{sec:PSD}). It constructs soft labels based on the similarity of local features, which solves the problems existing in traditional self-distillation methods on ReID. CSL and PSD are jointly trained in an end-to-end manner (Sec.~\ref{sec:training}).

\subsection{Transformer Encoder}
Our Transformer encoder $f$ consists of $L$ blocks. Each block contains global attention, part-aware attention and a feed-forward network. Global/Part-aware attention is used to extract global/local features. %After the input enters the attention layer, it will go through the same FFN layer output to obtain features of the same dimension as the input. 
%Now we introduce the input of Transformer encoder, global attention and part-aware attention respectively.
\label{sec:Part-ware Attention}
\paragraph{Input of Transformer Encoder.} We split an input image $x\in\mathbb{R}^{H \times W \times C}$ into non-overlapping $N$ patches by a patch embedding module. Each patch is treated as an ``image token" $\{x_i|i=1,2,\ldots,N\}$. Besides, a learnable ``class token" $x_{cls}$ and three ``part tokens"$\{x_{p_i}|i=1,2,3\}$ are concatenated with all ``image tokens". Then, the input to the Transformer encoder can be expressed as:
\begin{equation}
 \label{input}
     \mathcal{Z} = \left[ x_{cls}, x_{p_1}, x_{p_2}, x_{p_3}, x_1, \ldots, x_N\right] + \mathcal{P}
\end{equation}
where $\mathcal{Z}$ represents input sequence embeddings, 
$\mathcal{P} \in \mathbb{R}^{(N+4) \times D}$ is position embedding. $D$ is the number of channels.

The attention mechanism is based on a trainable associative memory with query $Q$, key $K$, and value $V$. They are all computed from the vector sequence $\mathcal{Z}$, which can be formulated as:
\begin{equation}
 \label{QKV}
  \begin{split}
       &Q = \mathcal{Z}W_Q = \left[ q_{cls}, q_{p_1}, q_{p_2}, q_{p_3}, q_1, \ldots, q_N\right]\\
       &K = \mathcal{Z}W_K = \left[ k_{cls}, k_{p_1}, k_{p_2}, k_{p_3}, k_1, \ldots, k_N\right]\\
       &V= \mathcal{Z}W_V  = \left[ v_{cls}, v_{p_1}, v_{p_2}, v_{p_3}, v_1, \ldots, v_N\right]
  \end{split}
\end{equation}
where $W_Q, W_K, W_V$ are different linear transformations.
%A block of the Transformer encoder contains layer norm, multi-head attention, and feed-forward layers. 

\paragraph{Global Attention.} To extract global features, we use ``class token" and all ``image tokens" to perform global attention. The output matrix of global attention can be obtained by:
\begin{equation}
\label{eq：self attn}
 Attention(\!Q_{cls}, \!K_{cls}, \!V_{cls}\!) \!\!=\!\! Softmax(\!\frac{Q_{cls}K_{cls}^T}{\sqrt{D}}\!)V_{cls}
\end{equation}
where $Q_{cls} = \left[ q_{cls}, q_1, \ldots, q_N\right]$, $K_{cls} = \left[ k_{cls}, k_1, \ldots, k_N\right]$ and $V_{cls}  = \left[ q_{cls}, q_1, \ldots, q_N\right]$. Then the outputs of global attention are sent to FFN network. Repeat this process $L$ times and we get a global feature $f(x_{cls})$.

\paragraph{Part-aware Attention.}To learn local similarity from data themselves, we need to extract local features using part-aware attention. For each ``part token" $x_{p_i}$, we use $x_{p_i}$ and ``image tokens" belonging to a special region to perform part-aware attention. the output matrix of part-ware attention can be obtained by:
\begin{equation}
 Attention(Q_{p_i}, K_{p_i}, V_{p_i}) \!\!=\!\! Softmax(\frac{Q_{p_i}K_{p_i}^T}{\sqrt{D}})V_{p_i}
\end{equation}
where $Q_{p_i} = \left[ q_{p_i}, q_{k_i}, \ldots, q_{k_i+m}\right]$ and $\{k_i,\ldots,k_i+m\}$ represent the serial number of $(m+1)$ ``image tokens" associated with the part token $x_{p_i}$. $Q_{p_i}$ and $K_{p_i}$ are handled in the same way. That is, ``part tokens" will only interact with ``image tokens" in the region of interest. In this work, We take three overlapping square areas along the vertical direction, as shown in the figure~\ref{fig:framework}. The outputs of part-ware attention are sent to the FFN network. Repeating this process $L$ times we get Three local features $\{f(x_{p_i})|i=1,2,3\}$. 
%They will be used in Cross-ID Similarity Learning~\ref{sec:CSL} and Part-guided Self-Distillation~\ref{sec:PSD}, respectively. 

\subsection{Cross-ID Similarity Learning}
\label{sec:CSL}
Existing Transformer-based ReID models perform representation learning based on global attention with annotated images, which leads the models to focus too much on domain-specific information, thus models get over-fitted on source domains. Besides, transformer-based ReID models ignore local similarities between different IDs, which can be helpful for generic representation learning. Specifically, cross-ID local similarities are ID-irrelevant, they offer visual knowledge to the models regardless of labels. Just like self-supervised learning, it provides the model with additional visual knowledge from the images themselves, not their labels.
% Supervised ReID models use hard labels for representation learning, which results in over-fitting. A common way to solve this problem is self-distillation, which provides the model with soft labels generated by the model itself to learn inter-class similarities. However, ReID is a fine-grained retrieval task, and the differences between IDs are not significant. So, it is difficult to generate useful soft labels from the model itself. As described in \ref{sec:CSL}, our CSL mines parts that are quite similar through different IDs. Based on the results of CSL, we propose part-.

To learn generic features, we propose a proxy task named Cross-ID Similarity Learning (CSL). Our method is based on the observation that although the entire images of different IDs are quite different, the local regions of some IDs are similar, such as red short sleeves, white skirts, and many more (see Figure~\ref{fig:intro_1}). The above information is not enough to discriminate ID on the source domain, but it is helpful to learn generic features. Just like other self-supervised learning methods in DG, such as solving jigsaw puzzles and predicting rotations, solving proxy tasks allow our model to learn generic features regardless of ReID task, and hence less over-fitting to domain-specific biases. 

Since it is difficult to find local similarities shared by different IDs in a mini-batch, we need to compare as many samples as possible in one gradient descent. To this end, we maintain a momentum-updated memory bank $\{w_{p_i}|i=1,\ldots,M\}$ for each part token $x_{p_i}$ during learning, which can be expressed as:
\begin{equation}
    \begin{split}
     w_{p_i}^{j} =\begin{cases}
        f\left( x_{p_i}^{j}\right) \,\,                                            &t=0,\\
        \left( 1-m \right) \times w_{p_i}^{j} +  m \times f\left( x_{p_i}^{j}\right)\,\,    &t>0\\
    \end{cases}
    \end{split}
\end{equation}
where $t$ is the training epoch, $m$ is the momentum and $j =1,\ldots,K $. $K$ is the number of samples in the source dataset.

For each local feature $f(x_{p_i}^j)$ in current mini-batch, we compare it with the entire memory bank. We select $k$ local features closest to $f(x_{p_i}^j)$ from $\{w_{p_i}\}^K$ to form a set $\{\mathcal{K}_{p_i}^j\}_{j=1}^k$ of positive samples. Then, the distance between positive samples and $f(x_{p_i}^j)$ is minimized by softmax-clustering loss
%to close the distance between positive samples and current features, and push other negative samples away. 
, which can be formulated as:
\begin{equation}
    \mathcal{L}_{p_i}^j=-log\frac{\sum_{w_{p_i}^m \in \{\mathcal{K}_{p_i}^j\}_{j=1}^k} \exp(\frac{f(x_{p_i}^j)w_{p_i}^m}{\tau})}{\sum_{n=1}^K \exp(\frac{f(x^j_{p_i})w_{p_i}^n}{\tau})}
\end{equation} 
where $\tau$ is a temperature coefficient. Minimizing $\mathcal{L}_{p_i}$ encourages the model to pull similar patches $\{\mathcal{K}_{p_i}^j\}_{j=1}^k$ close to $f(x_{p_i}^j)$ while pushing dissimilar patches away from $f(x_{p_i}^j)$ in feature space. In this way, the model can learn those visually similar patches in different IDs and make the Transformer notice the regions where this useful information is located.

%Note that ID information is not used in Apparent similarity Learning, but a self-supervised mode is used. As an auxiliary task, it focuses on learning the visual features of the image itself. This helps reduce the overfitting of the model on the source domain.

\subsection{Part-guided Self-Distillation}
\label{sec:PSD}
Self-distillation has been shown to help improve generalization. For example, it can learn visual similarities beyond hard labels and make the model converges easier to the flat minima~\cite{DG_SD,DG_SD2,SD_DG_vit}. However, our experiments found that the traditional self-distillation method could not improve the generalization of ReID. Because traditional self-distillation method relies on the output of the classifier to get the similarity between different categories. This is effective in image classification because of the large visual differences between different classes. For example, cats and dogs are visually similar, but cats and tables are very dissimilar. Such information can be utilized to facilitate learning. However, ReID is a fine-grained retrieval task, and the differences between different IDs are insignificant.
So there is no useful information in the classification results.

To this end, we propose Part-guided Self-Distillation (PSD), which uses the visual similarity of local parts to implement self-distillation. In Section \ref{sec:CSL}, each local representation $f(x_{p_i}^j)$ gets $k$ positive samples, and an image includes three part. Therefore, there are $3k$ positive samples in total for the global representation. We regard the IDs $\{I_i\}_{i=1}^{3k}$ corresponding to these $3k$ part as similar IDs. Soft labels $Y^j_s$ of $f(x_{cls}^j)$ are constructed as follows:
\begin{equation}
        \begin{split}
     Y_s^j|_i =\begin{cases}
        1-\alpha     &i=y_s^j\\
        \frac{\alpha}{3k}n_i                                          &i\in \{I_i\}_{i=1}^{3k},\\
        0        \,\,    &i\notin \{I_i\}_{i=1}^{3k}\cup \{y_s^j\}\\
    \end{cases}
    \end{split}
\end{equation} 
where $\alpha$ is the weight of similar categories, $n_i$ is the number of the i-th ID in $\{I_i\}_{i=1}^{3k}$ and $y_s^j$ is ground truth of $f(x_{cls}^j)$. That is, the more similar parts, the greater the probability of the ID to which these parts belong.

Then, the part-guided self-distillation loss can be formulated as:
\begin{equation}
    \mathcal{L}_s^j\!=\!-\!\lambda{Y_s^j\log \!P\!\left(\! f(x_{cls}^j\!)\right)}\!-\!(1\!-\!\lambda){Y^j\log P\left(\!f(x_{cls}^j\!)\right)}
\end{equation}
where $Y^j$ is the one-hot hard label, $\lambda$ is the coefficient to balance soft label and $P$ is the classifier that predicts probability distribution on source dataset. Since the apparent similarity is obtained by comparing the local representations of the current sample with the entire dataset. Therefore, it reflects the similarity between IDs better than the classification result of the classifier.
%There may be duplicate IDs in M*K patches. The more duplicate IDs, the more similar the current IDs are.

\subsection{Loss Function and Discussion}
\label{sec:training}
\paragraph{Loss function.} In addition to the softmax-clustering loss and part-guided self-distillation loss used in \ref{sec:CSL} and \ref{sec:PSD}, we also use the triplet loss with soft-margin to constrain the distance between positive and negative sample pairs. it can be formulated as follows:
\begin{equation}
    % \mathcal{L}_{tri}={[d_p - d_n + \alpha]_+}
    \mathcal{L}_{tri}={log[1 + exp({||{f}_{a} - {f}_{p}||}_{2}^{2} - {||{f}_{a} - {f}_{n}||}_{2}^{2})]}
\end{equation}
where \{$f_a$, $f_p$, $f_n$\} are the features of a triplet set.

For a mini-batch with $N_b$ samples, our entire loss function can be expressed as:
\begin{equation}
    \mathcal{L}= \mathcal{L}_{tri} + \sum_{j=1}^{N_b}{(\sum_{i=1}^M\mathcal{L}_{p_i}^j+{\mathcal{L}_s^j})}
\end{equation}
%\paragraph{Inference.} When the model migrates to unseen target domains for inference, we only use global attention in Equation~\ref{eq：self attn} to get a global representation for retrieval. Local representations are not used for inference but guide the model to mine discriminative information shared by different IDs during training. Therefore, our inference speed does not slow down as drastically as other local feature-based ReID models.

\paragraph{Discussion about occlusion and cloth-changing.} 
Since CSL and PSD mine the local similarity shared by different IDs, it may aggravate the negative impact of occlusion or cloth-changing. Below we briefly explain why our method is still valid. (1) We do not fuse local features (cascaded or weighted with global features) during inference, so local features with only occlusion information would not degrade performance. See Section 1 of the Appendix for details. (2) Since there is still a large gap between DG ReID and supervised ReID, the impact of cloth-changing can be ignored compared to improving generalization. 
%It makes sense to consider dressing changes only when the performance of the DG is close to the supervised level.

\section{Experiments}

%%%%%%%%%%% single-source
\begin{table*}[t]
  \centering
  \setlength{\abovecaptionskip}{0.cm}
  \caption{Performance comparisons between ours and the state-of-the-art in single-source DG ReID on Market1501, DukeMTMC-reID, MSMT17, and CUHK03-NP. Our results are highlighted in bold and others’ best results are underlined. The subscripts $_{50}$ and $_{152}$ denote using IBNNet$_{50}$ and IBNNet$_{152}$ as backbone, respectively.}
  \setlength{\tabcolsep}{3.8mm}
    \begin{tabular}{cc|c|cc|cc|cc}
    % \toprule
    % \toprule
    \midrule
    \midrule
    \multirow{2}[1]{*}{Method} & \multirow{2}[1]{*}{Reference} &  & \multicolumn{2}{c|}{Duke} & \multicolumn{2}{c|}{MSMT} & \multicolumn{2}{c}{CUHK-NP} \\
          &       &       & R1    & mAP   & R1    & mAP   & R1    & mAP \\
    % \midrule
    % \midrule
    % \midrule
    \midrule
    SNR \cite{SNR}   & CVPR2020 & \multirow{7}[1]{*}{Source:} & 55.1 & 33.6 & - & - & - & - \\
    QAConv$_{50}$ \cite{QAConv} & \multirow{2}[1]{*}{ECCV2020} & \multirow{7}[1]{*}{Market} & 48.8 & 28.7 & 22.6 & 7.0 & 9.9 & 8.6  \\
    QAConv$_{152}$ \cite{QAConv} &   &  & 54.4 & 33.6 & 25.6 & 8.2 & 14.1 & 11.8 \\
    % CBN   & ECCV2020 &  & \textcolor[rgb]{0.357, 0.608, 0.835}{58.7} & \textcolor[rgb]{0.357, 0.608, 0.835}{38.2} & 25.3  & 9.5   & -  & - \\
    TransMatcher \cite{transmatcher} & NeurIPS2021 &  & - & - & \uline{47.3} & \uline{18.4} & \uline{22.2} & \uline{21.4} \\
    MetaBIN \cite{metabin} & CVPR2021 &  & 55.2 & 33.1 & - & - & - & - \\
    QAConv-GS \cite{GS} & CVPR2022 &  & - & - & 45.9 & 17.2 & 19.1 & 18.1 \\
    MDA \cite{MDA}  & CVPR2022 &  & 56.7 & 34.4 & 33.5 & 11.8 & - & - \\
    % \midrule
    %TransReID-B/16 \cite{transreid} & baseline &  & 65.9 & 46.5 & 38.6 & 16.2 & 23.1 & 23.6 \\
    DTIN-Net\cite{DTIN} & ECCV2022 &  & \uline{57.0} & \uline{36.1} & - & - & - & - \\
    PAT & Ours &  & \textbf{67.9} & \textbf{48.9} & \textbf{42.8} & \textbf{18.2} & \textbf{25.4} & \textbf{26.0} \\
    % \midrule
    % \midrule
    \midrule

    \multirow{2}[1]{*}{Method} & \multirow{2}[1]{*}{Reference} &  & \multicolumn{2}{c|}{Market} & \multicolumn{2}{c|}{MSMT} & \multicolumn{2}{c}{CUHK-NP} \\
          &       &       & R1    & mAP   & R1    & mAP   & R1    & mAP \\
    % \midrule
    % \midrule
    \midrule
    % \midrule
    SNR \cite{SNR}   & CVPR2020 & \multirow{5}[1]{*}{Source:} & 66.7  & 33.9  & - & - & - & - \\
    % CBN   & ECCV2020 &       & \textcolor[rgb]{1.000, 0.000, 0.000}{72.7} & \textcolor[rgb]{1.000, 0.000, 0.000}{43.0} & \textcolor[rgb]{0.357, 0.608, 0.835}{35.4} & \textcolor[rgb]{0.357, 0.608, 0.835}{13.0} & - & - \\
    QAConv$_{50}$ \cite{QAConv} & \multirow{2}[1]{*}{ECCV2020} & \multirow{5}[1]{*}{Duke} & 58.6 & 27.2 & 29.0 & 8.9 & 7.9 & 6.8 \\
    QAConv$_{152}$ \cite{QAConv} &  &  & 62.8 & 31.6 & 32.7 & 10.4 & \uline{11.0} & \uline{9.4} \\
    % TransMatcher & NeurIPS2021 &  & 77.0* & 45.9* & - & - & 18.4* & 17.6* \\
    MetaBIN \cite{metabin} & CVPR2021 &       & 69.2  & 35.9  & - & - & - & - \\
    MDA \cite{MDA}  & CVPR2022 &       & \uline{70.3}  & \uline{38.0}  & \uline{39.8} & \uline{13.6} & - & - \\
    % \midrule
    %TransReID-B/16 \cite{transreid} & baseline &       & 68.9  & 42.6  & 40.2  & 16.7  & 17.9  & 18.6 \\
    DTIN-Net\cite{DTIN} & ECCV2022 &  & 69.8 & 37.4 & - & - & - & - \\
    PAT & Ours  &       & \textbf{71.9} & \textbf{45.2} & \textbf{43.9} & \textbf{19.2} & \textbf{18.8} & \textbf{18.9} \\
    % \midrule
    % \midrule
    \midrule
    \midrule
    %\multirow{2}[1]{*}{Method} & \multirow{2}[1]{*}{Reference} &  & \multicolumn{2}{c|}{Market} & \multicolumn{2}{c|}{Duke} & \multicolumn{2}{c}{CUHK-NP} \\
     %     &       &       & R1    & mAP   & R1    & mAP   & R1    & mAP \\
    % \midrule
    % \midrule
    % \midrule
    %\midrule
    % SNR   & CVPR2020 & \multirow{8}[1]{*}{MSMT} & 70.1  & 41.4  & 69.2 & 49.9 & - & - \\
    % CBN   & ECCV2020 &       & 73.7 & 45.0 & \textcolor[rgb]{0.357, 0.608, 0.835}{66.2} & \textcolor[rgb]{0.357, 0.608, 0.835}{46.7} & - & - \\
    %QAConv$_{50}$ \cite{QAConv} & \multirow{2}[1]{*}{ECCV2020} & \multirow{5}[1]{*}{Source:} & 72.6 & 43.1 & 69.4 & 52.6 & 25.3 & 22.6 \\
    %QAConv$_{152}$ \cite{QAConv} &  & \multirow{5}[1]{*}{MSMT} & 73.9 & 46.6 & \uline{72.2} & \uline{53.4} & 32.6 & 28.1 \\
    %TransMatcher \cite{transmatcher} & NeurIPS2021 &   & \uline{80.1} & \uline{52.0} & - & - & \uline{23.7} & \uline{22.5} \\
    %QAConv-GS \cite{GS} & CVPR2022 &   & 79.1 & 49.5 & - & - & 20.9 & 20.6 \\
    % MDA   & CVPR2022 &  & 79.7 & 53.0 & 71.7 & 52.4 & - & - \\
    % \midrule
    %TransReID-B/16 \cite{transreid} & baseline &       & 71.6 & 44.4 & 70 & 52.5 & 23.8 & 23.5 \\
    %PAT & Ours  &       & \textbf{72.2} & \textbf{47.3} & \textbf{70.7} & \textbf{53.8} & \textbf{24.2} & \textbf{25.1} \\
    % \bottomrule
    % \bottomrule
    % \midrule
    %\midrule
    \end{tabular}%
  \label{tab:single-source}%
\end{table*}%

\subsection{Datasets and Evaluation Metrics} \label{sec:4.1}
As shown in Table \ref{tab:Datasets}, we conduct experiments on four large-scale person re-identification datasets: Market1501\cite{Market}, DukeMTMC-reID\cite{Duke}, MSMT17\cite{MSMT} and CUHK03-NP\cite{cuhk03-np}. For simplicity, we denote the datasets above as M, D, MS, and C, respectively. We adopt the detected subset of the new protocol of CUHK03\cite{cuhk03} (767 IDs for training and 700 IDs for evaluation), which is more challenging than the original CUHK03 protocol. 
%%%%%%%% datasets intro
%%%%%%%%%%% dataset
\begin{table}[!h]
    \centering
    \setlength{\abovecaptionskip}{0.cm}
    \caption{Statistics of Person ReID Datasets.}
    \begin{tabular}{c|ccc}
        \midrule
        Dataset & \# IDs & \# images & \# cameras \\
        \midrule
        Market1501\cite{Market} & 1,501 & 32,217 & 6 \\
        DukeMTMC-reID\cite{Duke} & 1,812 & 36,411 & 8 \\
        MSMT17\cite{MSMT} & 4,101 & 126,441 & 15 \\
        CUHK03-NP\cite{cuhk03-np} & 1,467 & 28,192 & 2 \\
        \midrule
    \end{tabular}
    \label{tab:Datasets}
\end{table}

To evaluate the generalization of our models, we adopt a single-source protocol \cite{QAConv} and a multi-source protocol \cite{M3L}. Under the setting of single-source, we use one dataset mentioned above for training (only the training set) and another one for testing (only the testing set). Under the multi-source protocol, we select one domain from M+D+C3+MS for testing(only the testing set in this domain) and all the remaining domains for training (only the training set). For evaluation metrics, the performance is evaluated quantitatively by mean average precision (mAP) and cumulative matching characteristic (CMC) at Rank-1 (R1).

\subsection{Implementation Details}
We use ViT-base with $stride=16$ \cite{vit} pre-trained on ImageNet as our backbone (denoted as ViT-B/16 for short). The batch size is set to 64 and images are resized to $256\times128$. We adopt random flipping and local grayscale transformation \cite{lgt} for data augmentation. To optimize the model, we use SGD optimizer with a weight decay of $10^{-4}$. The learning rate increases linearly from 0 to $10^{-3}$ in the first 10 epochs then it decays in the following 50 epochs. The total training stage takes 60 epochs. For hyper-parameters, we conduct comprehensive experiments on the temperature parameter $\tau$ in section \ref{sec:ablation3}. Unless otherwise specified, we set $\alpha$ (the weight of similar categories), $\lambda$ (the coefficient to balance soft label and hard label), and $k$ (the number of clusters) to 0.5, 0.5, and 10, respectively. Besides, the label-smoothing parameter is 0.1. 
% We freeze the parameters of the patch embedding for more stable training.
As for baseline, we use TransReID-B/16 \cite{transreid} without SIE and JPM for a fair comparison. 
% TransReID-B/16 is pre-trained on ImageNet, then trained on source datasets, just like our model.

%%%%%%%%% multi-source(M3L)
\begin{table*}[]
    \centering
    \setlength{\abovecaptionskip}{0.cm}
    \caption{Comparison with the State-of-the-Arts DG models on four large-scale person ReID benchmarks: Market-1501 (M), DukeMTMC-reID (D), MSMT17 (MS) and CUHK03 (C). Our results are highlighted in bold and others’ best results are underlined.}
    \setlength{\tabcolsep}{1.2mm}
    % \begin{threeparttable}
    \begin{tabular}{c|c|cc|c|cc|c|cc|c|cc}
        \midrule
         \multirow{2}[1]{*}{Method} &  & \multicolumn{2}{c|}{MSMT} &  & \multicolumn{2}{c|}{Duke} & & \multicolumn{2}{c|}{Market} & & \multicolumn{2}{c}{CUHK} \\
         &  & R1 & mAP &  & R1 & mAP & & R1 & mAP & & R1 & mAP \\
         \midrule
         QACONV$_{50}$* \cite{QAConv} & \multirow{3}[1]{*}{Source:} & 29.9 & 10.0 & \multirow{3}[1]{*}{Source:} & 64.9 & 43.4 & \multirow{3}[1]{*}{Source:} & 68.6 & 39.5 & \multirow{3}[1]{*}{Source:} & 22.9 & 19.2 \\
         M$^3$L \cite{M3L} & \multirow{3}[1]{*}{M+D+C} & 33.0 & 12.9 & \multirow{3}[1]{*}{MS+M+C} & \uline{69.4} & 50.5 & \multirow{3}[1]{*}{MS+D+C} & 74.5 & 48.1 & \multirow{3}[1]{*}{MS+M+D} & 30.7 & 29.9 \\
        
         M$^3$L$_{IBN}$ \cite{M3L} &  & \uline{36.9} & \uline{14.7} & & 69.2 & \uline{51.1} & & \uline{75.9} & \uline{50.2} & & \uline{33.1} & \uline{32.1} \\
         % ViT-B/16 & & 76.0 & 52.1 & & 72.8 & 57.2 & & 45.7 & 22.0 & & 30.6 & 31.2 \\
         %M$^3$L(ViT-Base) \cite{M3L} & & 34.8 & 15.2 & & 71.6 & 55.8 & & 76.4 & 53.5 & & - & -  \\
         PAT & & \textbf{45.6} & \textbf{21.6} & & \textbf{71.8} & \textbf{56.5} & & \textbf{75.2} & \textbf{51.7} & & \textbf{31.1} & \textbf{31.5} \\
         \midrule
    \end{tabular}
    \label{tab:multi-source}%
    \vspace{-0.4em}
    \begin{tablenotes}
        \footnotesize
        \item * represents the results reimplemented by M$^3$L.
    \end{tablenotes}
    % \end{threeparttable}
\end{table*}

%%%%%%%%% ablation1.5 attention map visualization
\begin{table}[t]
    \centering
    \caption{Improvements of our method (PAT) on different Transformers. The training set is Market.}
    \resizebox{\columnwidth}{!}{
    \begin{tabular}{c|cc|cc|cc}
         \toprule
         \multirow{2}[1]{*}{Method}  & \multicolumn{2}{c|}{Duke} & \multicolumn{2}{c|}{CUHK} & \multicolumn{2}{c}{MSMT} \\
           & R1 & mAP  & R1  & mAP & R1  & mAP \\
         \midrule
         \midrule
         DeiT-Tiny\cite{DeiT} & 39.0 & 23.4 & 9.1 & 9.5 & 14.2  & 4.5 \\
         PAT & \textbf{48.5}  & \textbf{29.9} & \textbf{13.5}  & \textbf{14.2} & \textbf{19.2}  & \textbf{7.0} \\
         \midrule
         DeiT-Small\cite{DeiT}  & 53.0  & 34.0 & 14.1  & 14.5 & 23.5  & 8.4 \\
         PAT  & \textbf{56.9}  & \textbf{37.4} & \textbf{18.4}  & \textbf{18.2} & \textbf{27.1}  & \textbf{10.0} \\
         \midrule
         ViT-Small\cite{vit} & 51.9  & 33.1 & 12.9  & 14.0 & 24.8  & 9.2 \\
         PAT & \textbf{54.4}  & \textbf{35.9} & \textbf{15.1}  & \textbf{15.2} & \textbf{26.9}  & \textbf{10.0} \\
         \midrule
         DeiT-base\cite{DeiT}  & 59.3  & 41.3 & 18.1  & 18.8 & 32.9  & 12.8 \\
         PAT & \textbf{61.3}  & \textbf{42.6} & \textbf{20.3}  & \textbf{21.0} & \textbf{36.2}  & \textbf{14.9} \\
         \midrule
         ViT-Base\cite{vit}   & 65.9  & 46.5 & 23.1  & 23.6 & 38.6  & 16.2 \\
         PAT & \textbf{67.9}  & \textbf{48.9} & \textbf{25.4}  &\textbf{26.0} & \textbf{42.8}  & \textbf{18.2} \\

         \bottomrule
    \end{tabular}
    }
    \label{tab:transformer}
\end{table}

\subsection{Comparison with State-of-the-art Methods}
\paragraph{Single-source DG ReID} To validate the performance of our model, we evaluate our framework on the single-source generalization ReID benchmark. Specifically, we use Market-train and Duke-train as the training sets, and use Market-test, Duke-test, MSMT-test, and CUHK03-test as the testing sets. %In other words, only the training set of the source datasets is used.

The experimental results are shown in Table \ref{tab:single-source}. Our model outperforms the SOTA model under most settings and achieves a comparable performance with the SOTA model under the rest settings. In particular, under M$\to$D and D$\to$M settings, our model's mAPs outperform the state-of-the-art model by 10.3\% on average. 
When trained on Market, our model surpasses the SOTA model (DTIN-NET\cite{DTIN}) by 10.9\% and 12.8\% (test on Duke), 3.2\% and 4.6\% (test on CUHK03-NP) in R1 and mAP.
When trained on Duke, our model outperforms the SOTA by 1.6\% and 7.2\% (test on Market), 4.1\% and 5.6\% (test on MSMT) for R1 and mAP, respectively. 
This demonstrates the superiority of our model. 
%When trained on MSMT, our model still achieves SOTA on CUHK03-NP. 
%But TransMatcher \cite{transmatcher} surpasses our method on Market. 
%Besides, we experimentally found that the IBN \cite{BINnet,ibn2} trick greatly improved it, and our model could still surpass it if TransMatcher used a normal CNN as the backbone. This result is shown in the Appendix.

%Notably, our model achieves a decent performance even just trained on a small dataset: under the Market$\to$Duke setting, the performance is only lower 4.9\% than under the MSMT$\to$Duke setting in mAP. It indicates that our model is good at dealing with over-fitting caused by the limitation of the data volume. 

\paragraph{Multi-source DG ReID} To further validate the generalization of our model, we also conduct the experiments under the multi-source protocol. We follow the protocol proposed in M$^3$L \cite{M3L} which we have introduced in section \ref{sec:4.1}. We use three of Market1501 (M), DukeMTMC-reID (D), MSMT17 (MS) and CUHK03-NP (C) as our source domain and the rest as our target domain. Note that, we follow the performance of QAConv reimplemented by M$^3$L.

%When conducting MS+M+C$\to$D or M+D+C$\to$MS, 
As shown in Table \ref{tab:multi-source}, our method outperforms the SOTA model on all datasets. Specifically, our model's mAP is 6.9\% higher than the current best model (M$^3$L) under the M+D+C$\to$MS setting and reaches 56.5\% under the MS+M+C$\to$D setting, which exceeds M$^3$L by 5.4\%. When training on MS+D+C or MS+M+D, our model still shows quite a similar performance compared with M$^3$L.

\begin{table}[]
    \centering
    \setlength{\abovecaptionskip}{0.cm}
    \caption{Ablation studies on the effectiveness of CSL and PSD. All methods are trained on Market and evaluated on Duke, MSMT, and CUHK03-NP.}
    \setlength{\tabcolsep}{1.mm}
    %\resizebox{0.8\columnwidth}{!}{
    \begin{tabular}{c|cc|cc}
    \midrule
         \multirow{2}[1]{*}{Method} & \multicolumn{2}{c|}{M$\to$D} & \multicolumn{2}{c}{M$\to$MS}\\ %& \multicolumn{2}{c}{M$\to$C} \\
         & R1 & mAP & R1 & mAP \\ %& R1 & mAP 
         \midrule
         \midrule
         Baseline (B) & 65.9 & 46.5 & 38.6 & 16.2\\ %& 23.1 & 23.6 
         \midrule
         B+SD \cite{SD_DG_vit} & 64.1 & 44.9 & 35.6 & 14.6 \\%& 21.3 & 21.8 \\
         %Baseline + hint & 66.3 & 47.3 & 40.7 & 17.1 & 25.4 & 24.6 \\
         %\midrule
         B+CSL & 67.1 & 48.3 & 40.2 & 17.0\\ % & 25.4 & 25.9 \\
         B+CSL+PSD (ours)  & \textbf{67.9} & \textbf{48.9} & \textbf{42.8} & \textbf{18.2}\\ % & \textbf{25.4} & \textbf{26.0} \\
         \midrule
    \end{tabular}
    %}
    \label{tab:ablation1}
\end{table}

%%%%%%%%% ablation
\subsection{Ablation Study}
\paragraph{Improvement on Transformer.}
To investigate the improvement of our method on original Transformers (baseline), we conduct extensive experiments on different Transformers. As shown in Tab \ref{tab:transformer}, our approach can improve the generalization of various Transformers on ReID task, especially when model size is small. For example, our method surpasses baseline by 9.5\% and 6.5\% in R1 and mAP (Market $\to$ Duke) when using DeiT-Tiny as the backbone.

\paragraph{Ablation study of main components of our model.}
To ensure that all components promote our model, we conduct an ablation study. All models are trained on Market and then tested on Duke, MSMT, and CUHK03-NP, respectively. We choose the following models: (1) Baseline, namely TransReID-B/16 without SIE and JPM which is introduced in \ref{sec:4.1}; (2) Baseline with conventional self-distillation (B + SD). We follow the self-distillation way designed for domain generalization \cite{SD_DG_vit}; (3) our model without PSD (B + CSL); (4) our model with all components, including CSL and PSD.
%(3) Baseline with hint learning (Baseline + hint). We drop the head layers for self-distillation of (2) and turn to pull normalized features of intermediate blocks and the final block; 

As shown in Table \ref{tab:ablation1}, Baseline + SD (self-distillation) results in a descent, which means that traditional self-distillation fails to improve the generalization of ReID. Since ReID is a fine-grained retrieval task, it is not suitable to use conventional self-distillation which usually requires large inter-class distances. The results of (3) and (4) demonstrate our contribution. Firstly, our CSL module brings significant improvement. Secondly, PSD using the visual similarity of local parts to construct soft labels is effective.
%On the whole, our model digs local similarities across different IDs to learn a more generic representation, that boosts generalization.
%and then uses it to perform PSD, which alleviates the model from taking a shortcut.

%%%%%%%%% ablation2 attention map visualization
\begin{figure}[t]
    \centering
    \setlength{\abovecaptionskip}{0.cm}
    \begin{center}
        \includegraphics[width=\columnwidth]{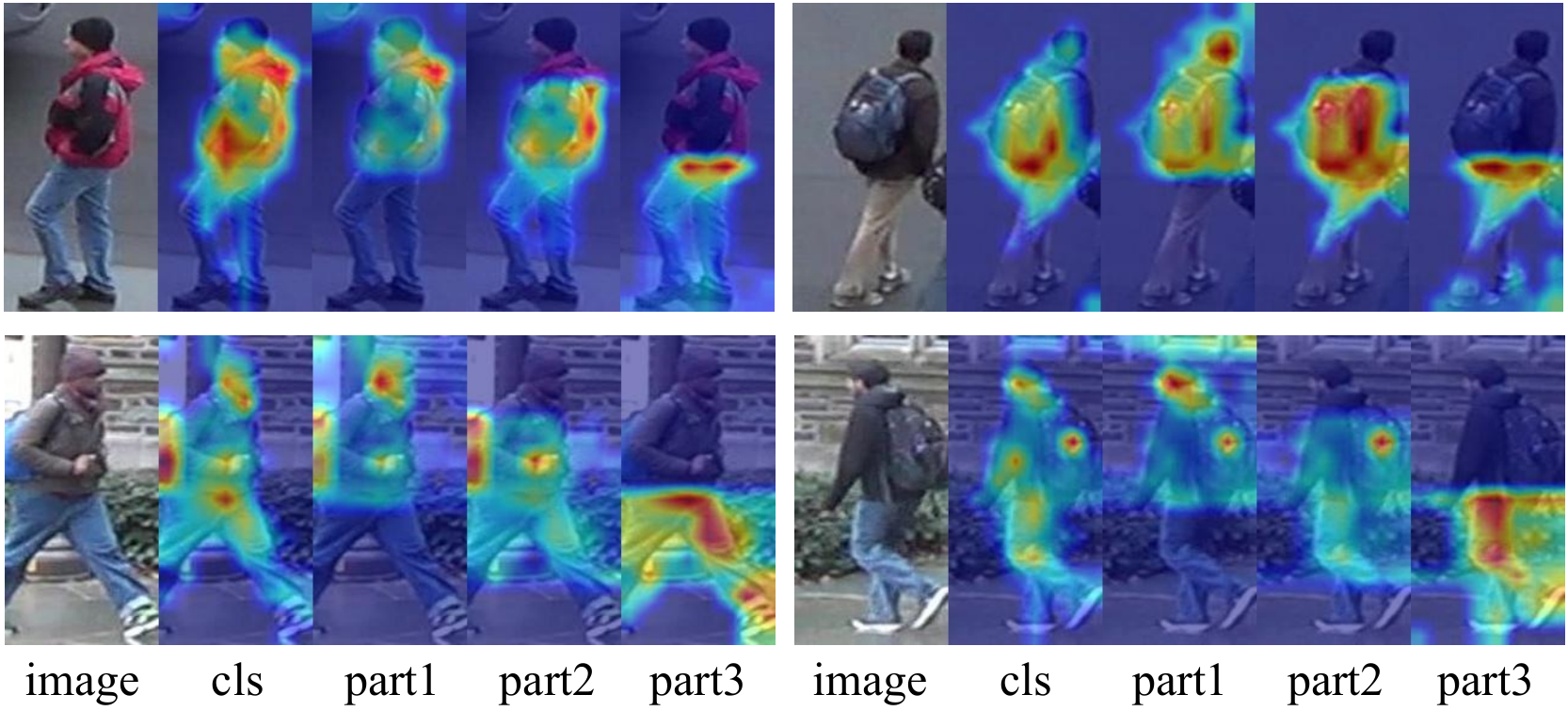}
    \end{center}
    \caption{Attention map visualization. Original images are selected from the target domain (Duke). We exhibit the visualizations of the class token and three part tokens.}
    \label{fig:ablation2}
\end{figure}

\begin{figure}[]
    \centering
    \setlength{\abovecaptionskip}{0.cm}
    \begin{center}
        \includegraphics[width=\columnwidth]{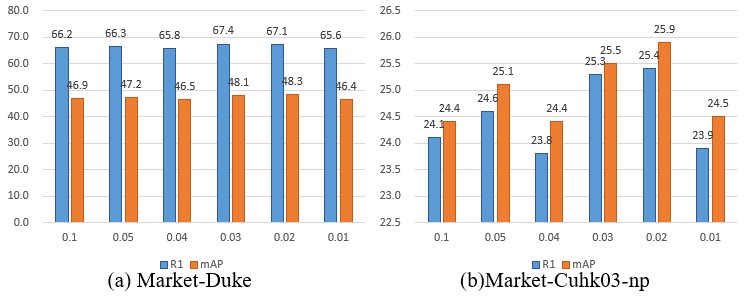}
    \end{center}
    \caption{Ablation study on $\tau$ of softmax-clustering loss.}
    \label{fig:ablation3}
\end{figure}

\paragraph{Visualization of attention maps.}
To better understand the part tokens, we conduct illustrative experiments on visualizing the attention maps. The size of attention maps in each head is N$\times$N (N is the number of patches). We obtain the first four rows of an attention map, which denote the attention on each image patch of the class token and the part tokens. We resize these four attentions to H$\times$W (H and W denote the height and width of the input), then we turn them into heat maps with the original image. We fuse the attention results of the shallow layers, which contain more visual information than the deep ones.

As shown in Figure \ref{fig:ablation2}, the class token mainly focuses on the whole images, while the ``part tokens" pay attention to local areas like the upper body, legs, and backpacks. It shows that ``part tokens" broaden the scope of attention, and provide more comprehensive multi-view information to the class token. To put it another way, our model learns a more generic representation by utilizing local similarities.

\label{sec:ablation3}
\paragraph{Ablation study on hyperparameters.}
%To investigate the effects of the temperature factor $\tau$ of softmax-clustering loss, we conduct experiments trained on Market with $\tau$ ranging from 0.01 to 0.1. 
(1) $\tau$: As shown in Figure \ref{fig:ablation3}, it is apparent that when trained on Market, setting $\tau$ to 0.02 is optimum. Softmax-clustering loss raises when $\tau$ decreases. If we set $\tau$ too large, the effect of softmax-clustering loss will be inappreciable. On the contrary, if $\tau$ is too small, the training stage will become quite unstable due to the increasing effect of softmax-clustering loss. Based on our experimental observation, we hold the opinion that using a larger $\tau$ is more suitable when the scale of the training set increases. 
(2) Number of parts: Dividing into three parts is the most intuitive and the most effective. The results of other numbers are shown in Table \ref{tab:part_num}. For analysis of other hyperparameters (such as $\alpha$, $\lambda$), see Appendix.
\begin{table}[]
    \centering
    \setlength\tabcolsep{5pt}
    \renewcommand{\arraystretch}{0.9} % Default value: 1
    \setlength{\abovecaptionskip}{0.cm}
    \caption{Ablation study for the number of parts.}
    \small
    \begin{tabular}{c|cc|cc|cc}
    
         \hline
         \multirow{2}[1]{*}{Number of parts} & \multicolumn{2}{c|}{M$\to$D} & \multicolumn{2}{c|}{M$\to$C} & \multicolumn{2}{c}{M$\to$MS}  \\
         & mAP & R1 & mAP & R1 & mAP & R1 \\
         \hline
         2 & 47.0 & 66.7 & 24.5 & 23.6 & 16.7 & 39.3 \\
         3 & 48.9 & 67.9 & 26.0 & 25.4 & 18.2 & 42.8 \\
         4 & 48.6 & 67.1 & 25.1 & 25.1 & 18.1 & 41.7 \\
         5 & 47.9 & 66.1 & 25.0 & 25.6 & 17.0 & 40.5 \\
         \hline
    \end{tabular}
    \label{tab:part_num}
\end{table}

\begin{figure}[]
    \centering
    \setlength{\abovecaptionskip}{0.cm}
    \begin{center}
        \includegraphics[width=0.9\columnwidth]{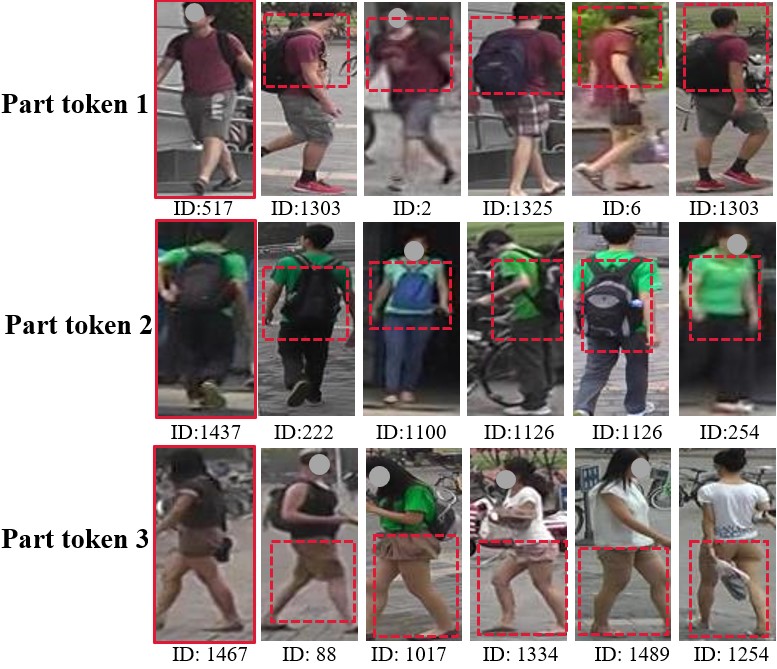}
    \end{center}
    \caption{Visualization of local features' ranking list in CSL.}
    \label{fig:ablation4}
\end{figure}

\paragraph{Visualization of local features' ranking list.} To verify whether CSL has mined local similarities, we show those samples that are closest to the current local feature, that is, the samples belonging to $\{\mathcal{K}_{p_i}^j\}_{j=1}^k$ in Section~\ref{sec:CSL}. As shown in Figure~\ref{fig:ablation4}, the model can find samples with apparent similarity to the current sample in a specific area. For example, in the area attended by part token 1, not only people wearing dark red shirts but also backpacks were found. The above similarity is not dependent on the labels but completely derived from the data itself, which guides the model to learn generic features. %CSL uses this similarity to learn visual information shared by different IDs. It helps the model pay attention to more discriminative information and extract more generalizable features.

\section{Conclusion}
In this paper, we propose a pure Transformer-based framework (termed Part-aware transformer) for DG ReID for the first time. Specifically, we design a proxy task, named Cross-ID Similarity Learning (CSL), to mine local visual information shared by different IDs. This proxy task allows the model to learn generic features because it only cares about the visual similarity of the parts regardless of the ID labels, thus alleviating the side effect of domain-specific biases. Furthermore, we propose a part-guided self-distillation module to further improve the generalization of the global representation. Experimental results on multi-source and single-source DG ReID show that our method achieves state-of-the-art in DG ReID.
% \newpage
%\paragraph{Broader impacts} The most significant contribution of ReID is to improve the accuracy of automatic person recognition, autonomous driving, and other fields. Our work effectively improves the accuracy of DG ReID, which makes the ReID system more practicable in security. However, ReID may bring privacy issues to our society. Firstly, ReID uses images that involve the privacy of pedestrians. These datasets should be carefully distributed and not used in illegal ways. Secondly, the ReID system may intentionally or unintentionally cause an invasion of privacy, so the deployment and application of the systems should be strictly controlled. To avoid privacy breaches due to face images, we only use the back and side views of pedestrians for display.

{\small
\bibliographystyle{ieee_fullname}
\bibliography{egpaper_final}
}

\end{document}